\documentclass[11pt]{article}

\usepackage[preprint]{acl}

\usepackage{times}
\usepackage{latexsym}
\usepackage[T1]{fontenc}
\usepackage[utf8]{inputenc}
\usepackage{microtype}
\usepackage{inconsolata}
\usepackage{multirow} 
\usepackage{amsmath,amsfonts,bm}

\def\eqref#1{equation~\ref{#1}}
\def\1{\bm{1}}

\DeclareMathAlphabet{\mathsfit}{\encodingdefault}{\sfdefault}{m}{sl}
\SetMathAlphabet{\mathsfit}{bold}{\encodingdefault}{\sfdefault}{bx}{n}

\usepackage{amssymb,amsthm,mathtools}
\usepackage{graphicx,booktabs,array,float,placeins}
\usepackage{algorithm,algpseudocode}
\usepackage{stfloats} % Allow wide floats at the bottom of a two-column page.
\usepackage{xurl}
\usepackage{needspace}
\graphicspath{{figures/}}
\newtheorem{proposition}{Proposition}
\newtheorem{theorem}{Theorem}
\newcommand{\method}{\textsc{CoS-VQ}}

\hypersetup{
  pdftitle={CoS-VQ: Coordinated Shared-Scale Vector Quantization for Large Language Models},
  pdfauthor={Zukang Xu, Zhixiong Zhao, Xing Hu, Zhe Jiang, Xin Si, Dawei Yang},
  pdfsubject={Large language model vector quantization}
}

\title{\method{}: Coordinated Shared-Scale Vector Quantization for Large Language Models}
\author{
  Zukang Xu\textsuperscript{1}\thanks{Equal contribution.}\quad
  Zhixiong Zhao\textsuperscript{1,2}\footnotemark[1]\quad
  Xing Hu\textsuperscript{1,3}\\
  \textbf{Zhe Jiang\textsuperscript{3}}\quad
  \textbf{Xin Si\textsuperscript{3}}\quad
  \textbf{Dawei Yang\textsuperscript{1}}\thanks{Corresponding author: \texttt{dawei.yang@houmo.ai}.}\\
  \textsuperscript{1}Houmo AI\\
  \textsuperscript{2}Peking University\\
  \textsuperscript{3}Southeast University
}

\begin{document}
\maketitle
\begin{abstract}
Large language models (LLMs) require substantial memory, motivating low-bit weight quantization for efficient deployment.
Existing vector quantization (VQ) methods often improve accuracy by increasing the effective coding dimension, which can complicate independent lookup and local codeword updates.
We propose \method{} (Coordinated Shared-Scale Vector Quantization), which preserves low-dimensional codewords while jointly optimizing hard indices and their shared scale across positions.
The key challenge is scale-induced coupling: the optimal scale depends on the entire codeword combination and, in turn, changes candidate rankings across positions, so pointwise updates can stall even after scale re-optimization.
\method{} collects candidates at multiple probe scales and constructs a tangent separable upper bound for the residual-aware quadratic objective.
By reducing candidate comparisons to lower envelopes of lines over positive scales, it identifies finitely many representative intervals for joint index--scale updates without exhaustively enumerating codeword combinations.
On Qwen3.8-27B, \method{} outperforms QTIP and LLVQ on the evaluated language modeling, coding, and reasoning benchmarks.
Ablations from the same initialization further show larger gains from joint updates than from scale-only refinement, highlighting joint selection as an alternative to increasing codeword dimensionality.
The code is available at \url{https://github.com/xzktx003/Cos_VQ}.
\end{abstract}
\section{Introduction}

% LLM background and motivation.
Large language models (LLMs) have demonstrated remarkable
capabilities across diverse tasks, including natural language
understanding, complex reasoning, and code generation.
Despite these advances, the growing scale of LLMs poses
significant challenges to their efficient deployment.
Representative models such as LLaMA-3.1-70B
~\citep{llama3herdmodels} and Qwen3.8-27B
~\citep{qwen3.8} contain tens of billions of parameters,
with a 70B-parameter model alone requiring approximately
140 GB of memory to store its weights in 16-bit precision.
Beyond the substantial memory footprint, autoregressive
inference involves repeated weight access, placing
considerable pressure on memory bandwidth and further
limiting deployment efficiency.
These challenges have driven growing interest in model
compression techniques that reduce memory requirements
and inference costs while preserving the capabilities
of the original models.

% PTQ, VQ, and limitations of high-dimensional coding.
Post-training quantization (PTQ) has emerged as a practical
paradigm for reducing the storage and inference costs of LLMs
without full model retraining
~\citep{zhao2026specquant,zhao2026all,pmlr-v306-zhao26bi,zhao2026bwla,zhao2025quark}.
Among existing approaches, vector quantization (VQ) represents
multiple weights using a single code index, offering greater
representational flexibility than conventional scalar
quantization, particularly at extremely low bitrates.
Recent methods such as QTIP and LLVQ further improve
quantization quality by exploiting high-dimensional
representations through trellis and lattice structures
~\citep{qtip,llvq}. Although these structures alleviate the exponential search
and storage costs associated with unstructured high-dimensional
codebooks, they introduce structural dependencies that can
restrict the independent lookup and local modification of
small codewords. This raises a fundamental question:
\emph{Can low-dimensional codewords retain their independent
lookup and local update capabilities while achieving the
quantization quality of high-dimensional structured methods
through a broader range of joint decisions?}

\begin{figure*}[t]
    \centering
    \includegraphics[width=0.95\linewidth]{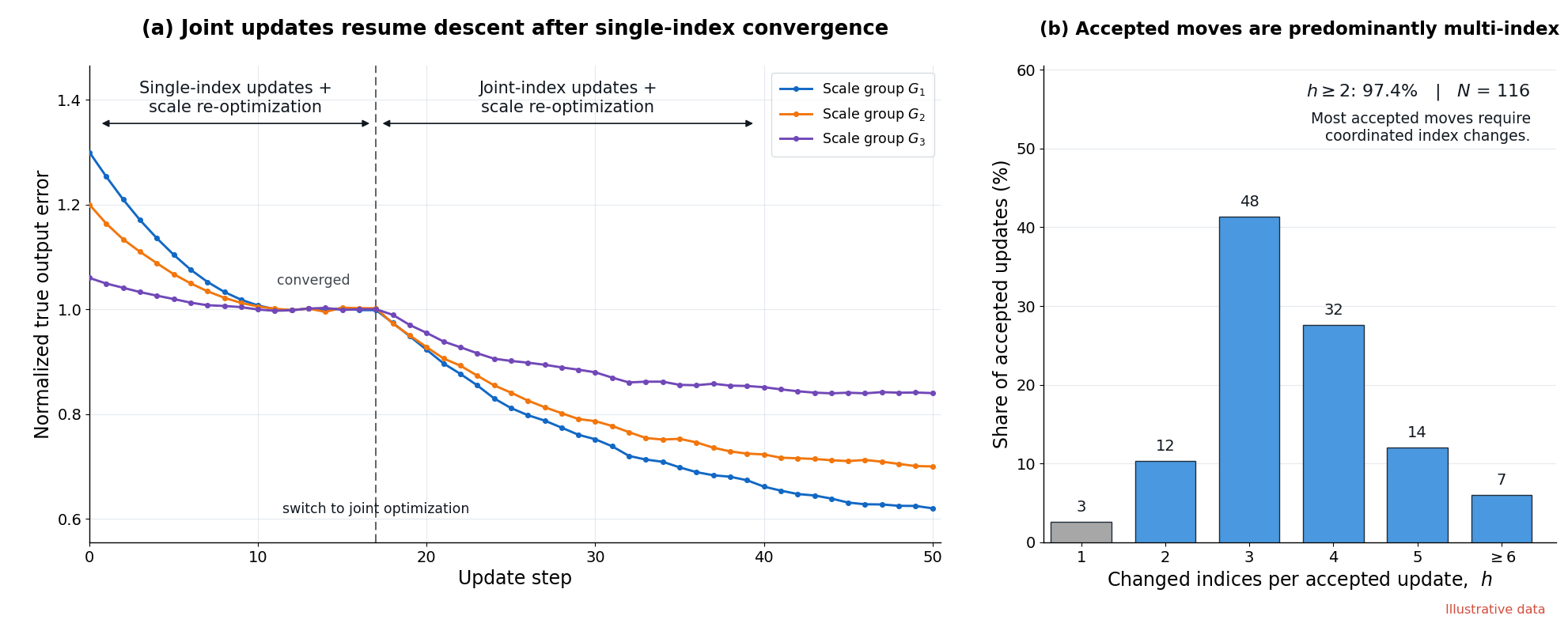}
    \caption{
    (a) Three shared-scale groups approach a coordinate-wise
    stall under single-index updates with scale re-optimization.
    After switching to joint-index updates, the normalized
    true layer-output error decreases further.
    Each curve is normalized to 1 at the switching point.
    (b) Distribution of the number of changed indices among
    accepted updates, showing that multi-index updates
    account for the majority.
    }
    \label{fig:motivation}
\end{figure*}

% Key observation and motivation.
To answer this question, we distinguish joint representation
within a vector from joint selection across multiple vectors.
In shared-scale VQ, multiple positions within a scale group
are assigned low-dimensional codewords while sharing a common
scale parameter.
Crucially, the optimal scale depends on the entire codeword
assignment and, in turn, changes the relative preference
among candidate codewords at different positions.
This bidirectional coupling creates a challenging optimization
landscape: even when the shared scale is re-optimized after
each single-index update, coordinate-wise refinement can
stall because no individual index change reduces the
reconstruction error, whereas coordinated changes across
multiple positions can still yield improvements.
As illustrated in Figure~\ref{fig:motivation}, the
reconstruction error plateaus under pointwise refinement
but decreases again after switching to joint-index updates,
with accepted updates frequently modifying multiple indices.
These observations motivate joint codeword selection under
a shared scale, enabling a broader optimization scope
without increasing the dimensionality of the underlying
codewords.

\begin{figure*}[t]
    \centering
    \includegraphics[width=0.8\linewidth]{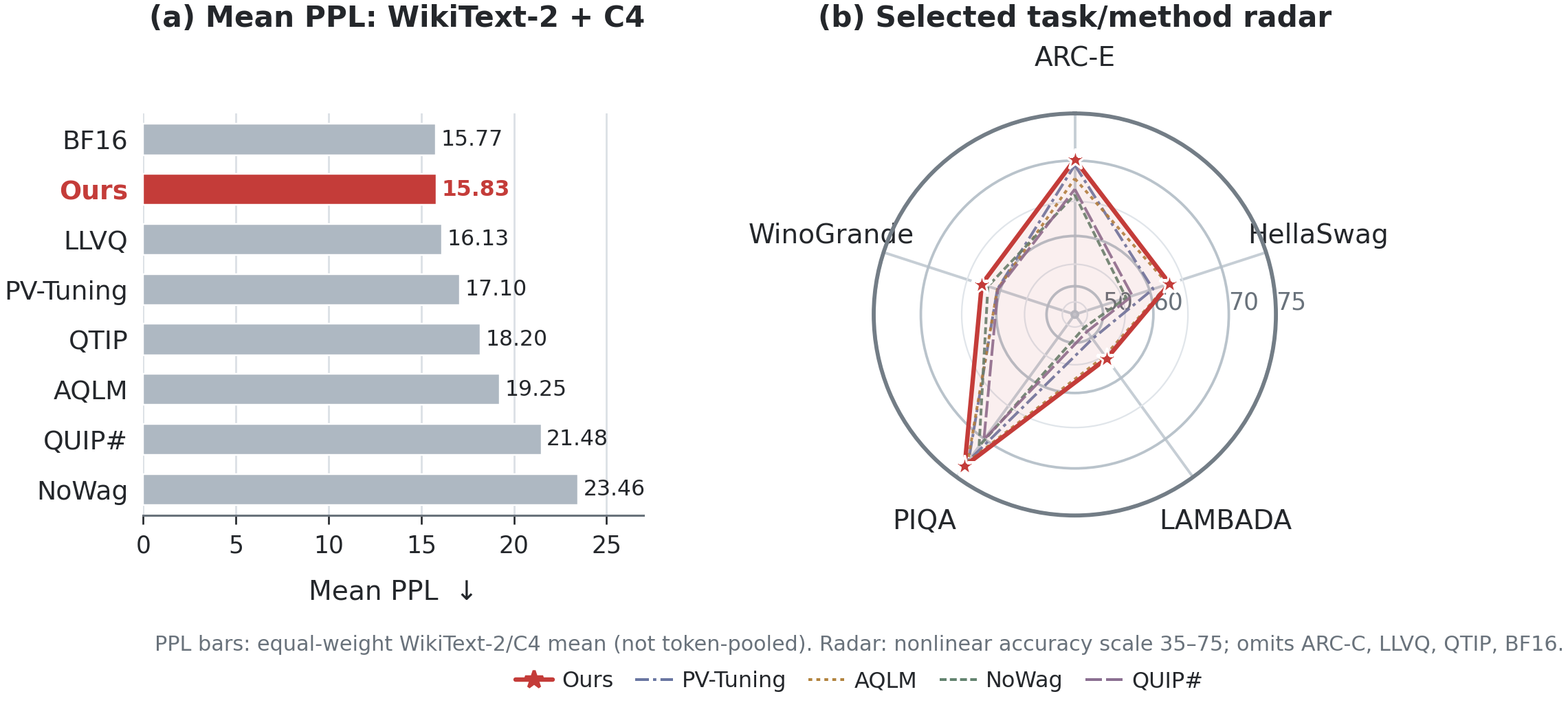}
    \caption{
    Overview of Qwen3-4B results.
(a) Equal-weight mean of WikiText-2 and C4 perplexity
across the evaluated low-bit quantization methods.
(b) Comparison of selected methods on five downstream tasks.
    }
    \label{fig:qwen3-4b-overview}
\end{figure*}

% Proposed method.
To address this challenge, we propose \textbf{\method{}}
(Coordinated Shared-Scale Vector Quantization), which
jointly optimizes hard codeword indices and their shared
scale while preserving low-dimensional codewords.
Directly optimizing these variables is computationally
intractable because the layer-output reconstruction
objective contains cross-position quadratic interactions,
and exhaustive enumeration of codeword combinations
incurs exponential complexity.
We first construct candidate pools by combining local
shortlists obtained at multiple probe scales, preventing
the search from being restricted to candidate rankings
under a single scale.
We then derive a residual-aware tangent separable upper
bound that transforms the coupled quadratic objective
into candidate-wise quadratic functions of the shared scale.
By eliminating candidate-independent terms, codeword
comparisons over positive scales can be expressed as
lower envelopes of lines.
Merging their switching points partitions the scale
domain into finitely many representative intervals,
within each of which the joint codeword assignment
remains constant and the scale can be optimized through
a constrained one-dimensional quadratic problem.
The resulting solver directly updates hard indices and
the shared scale without probabilistic relaxation or
gradient-based optimization, while verifying accepted
updates against the true layer-output error.
The experiments further support this joint-selection
design, as illustrated by the language-modeling and
downstream-task results in
Figure~\ref{fig:qwen3-4b-overview}.
Our key contributions can be summarized as follows:

\begin{itemize}
    \item We propose CoS-VQ, a coordinated shared-scale VQ
    framework that jointly optimizes hard indices and
    shared scales while preserving low-dimensional codewords.

    \item We develop a residual-aware separable upper bound
    and a lower-envelope interval solver, with theoretical
    guarantees on surrogate optimality and interval complexity.

    \item Extensive experiments across multiple LLMs
    demonstrate superior low-bit quantization quality,
    supported by joint-optimization and codeword-dimension
    ablations.
\end{itemize}

% 相关工作
\section{Related Work}

% 向量量化方法。
\paragraph{Vector quantization methods.}
Low-bit quantization methods span scalar quantization, explicit
vector codebooks, additive or residual codebooks, and structured
lattice or trellis encodings.
GPTVQ~\cite{gptvq} combines vector representations with layer
reconstruction; VPTQ~\cite{vptq} constructs explicit codebooks
using second-order objectives; AQLM~\cite{aqlm} uses additive
codebooks and block-level refinement; and CRVQ~\cite{crvq}
relaxes constraints for critical channels.
QuIP\#~\cite{quipsharp} combines incoherence processing with
lattice quantization, while QTIP~\cite{qtip} and LLVQ~\cite{llvq}
exploit trellis and lattice structures to increase effective
coding dimensionality.
OstQuant~\cite{ostquant} learns orthogonal and scaling
transforms for scalar grids.
RSAVQ~\cite{rsavq}, PCDVQ~\cite{pcdvq},
RWKVQuant~\cite{rwkvquant}, and KBVQ-MoE~\cite{kbvqmoe}
leverage curvature, polar magnitude--direction decoupling,
hybrid scalar--vector codes, and expert structure, respectively.
Unstructured high-dimensional codebooks incur costly
nearest-neighbor search and cache demands, whereas structured
encodings depend on constraints or specialized decoding kernels.
In contrast, we study low-dimensional codebooks with
a shared scale, retaining portability, independent lookup,
and local rewriting while using coordinated cross-position
selection to approach the quality of high-dimensional
structured quantization.

% 量化优化方法。
\paragraph{Quantization optimization methods.}
On interaction search and parameter updates, GPTVQ~\cite{gptvq} employs sequential compensation,
AQLM~\cite{aqlm} adopts beam search,
VPTQ~\cite{vptq} leverages second-order refinement,
CRVQ~\cite{crvq} models critical channels, and
KBVQ-MoE~\cite{kbvqmoe} accounts for expert-aggregation bias.
PV-Tuning~\cite{pvtuning}, GPTVQ~\cite{gptvq},
GSQ~\cite{gsq}, and LLVQ~\cite{llvq} explore
alternating updates or closed-form solutions for discrete
assignments and continuous scales.
In contrast, \method{} jointly optimizes hard indices
and their shared positive scale through scale elimination,
a residual-aware tangent separable upper bound, and
lower-envelope interval search, avoiding heuristic
enumeration of the Cartesian product.
\section{Preliminaries and Problem Formulation}
\label{sec:preliminaries}

\paragraph{Shared-scale vector quantization.}
Given a weight matrix $W\in\mathbb R^{p\times n}$ and calibration
inputs $X\in\mathbb R^{n\times N}$, we adopt fixed-codebook VQ.
We partition the weights into $d$-dimensional positions, where
position $i$ belongs to scale group $\mathcal G_g$ and is reconstructed as
\begin{equation}
\widehat w_i=s_g c_{k_i},\qquad c_{k_i}\in\mathcal C_i\subset\mathbb R^d.
\label{eq:recon}
\end{equation}
Positions may share a base codebook, with $\mathcal C_i$ denoting
the codebook accessible to position $i$.
We use $d=6$ by default and ablate it under matched actual bitwidth.
The joint selection range $m$, independent of $d$, specifies the
number of indices that may change per round.
The codebook, index bitwidth, scale-group partition, and scale
count remain fixed; only hard indices and existing scales are updated.
For a group of $M$ positions with $m\le M$ mutable indices,
the remaining $M-m$ positions enter the objective with singleton
candidate pools, since scale changes also affect their reconstructions.

\paragraph{Layer-output error.}
We minimize the calibration layer-output error
\begin{equation}
F(\widehat W)=\|WX-\widehat W X\|_F^2.
\label{eq:true_obj}
\end{equation}
Each round selects a complete scale group while fixing the
out-of-group state.
Let $z$ concatenate its weight vectors, with current value $z_0$,
and define a linear operator $\mathcal A$ mapping an in-group
modification $\delta$ to the vectorized output change after
scattering $\delta$ into the weight matrix.
With $r_0=\operatorname{vec}(WX-\widehat W_0X)$, the group
objective becomes $\|r_0-\mathcal A(z-z_0)\|_2^2$.
Each position has a finite candidate pool $\mathcal P_i$ containing
its current codeword, and $s\in\mathcal S=[s_{\min},s_{\max}]$,
where $0<s_{\min}\le s_{\max}<\infty$ and the current scale is feasible.
For floating-point storage, $\mathcal S$ is intersected with
the set of legally representable scales.

\paragraph{Joint coupling under a shared scale.}
Shared-scale coupling is evident even in the simpler Euclidean
objective $E(k,s)=\sum_i\|w_i-sc_{k_i}\|_2^2$.
When the denominator is nonzero and the solution is feasible,
the optimal scale is
\begin{equation}
s^*(k)=\frac{\sum_i w_i^\top c_{k_i}}{\sum_i\|c_{k_i}\|_2^2}.
\label{eq:simple_scale}
\end{equation}
At a fixed scale, position-wise selection is separable;
re-optimizing the scale couples candidate choices through
the entire codeword assignment.
Eq.~\ref{eq:simple_scale} illustrates this mechanism but does
not generally solve Eq.~\ref{eq:true_obj}.
The true output error also involves quadratic cross-position
interactions induced by calibration inputs and position layout.
We therefore construct a residual-aware surrogate and exploit
the shared-scale structure for optimization.
Each round activates at most four existing scale groups per row
(\texttt{groups\_per\_row=4}), a search budget that introduces
no additional stored scales; within each group, $m$ limits
the number of mutable indices.
\section{Method}
\label{sec:method}

\begin{figure*}[!t]
\centering
\includegraphics[width=\linewidth]{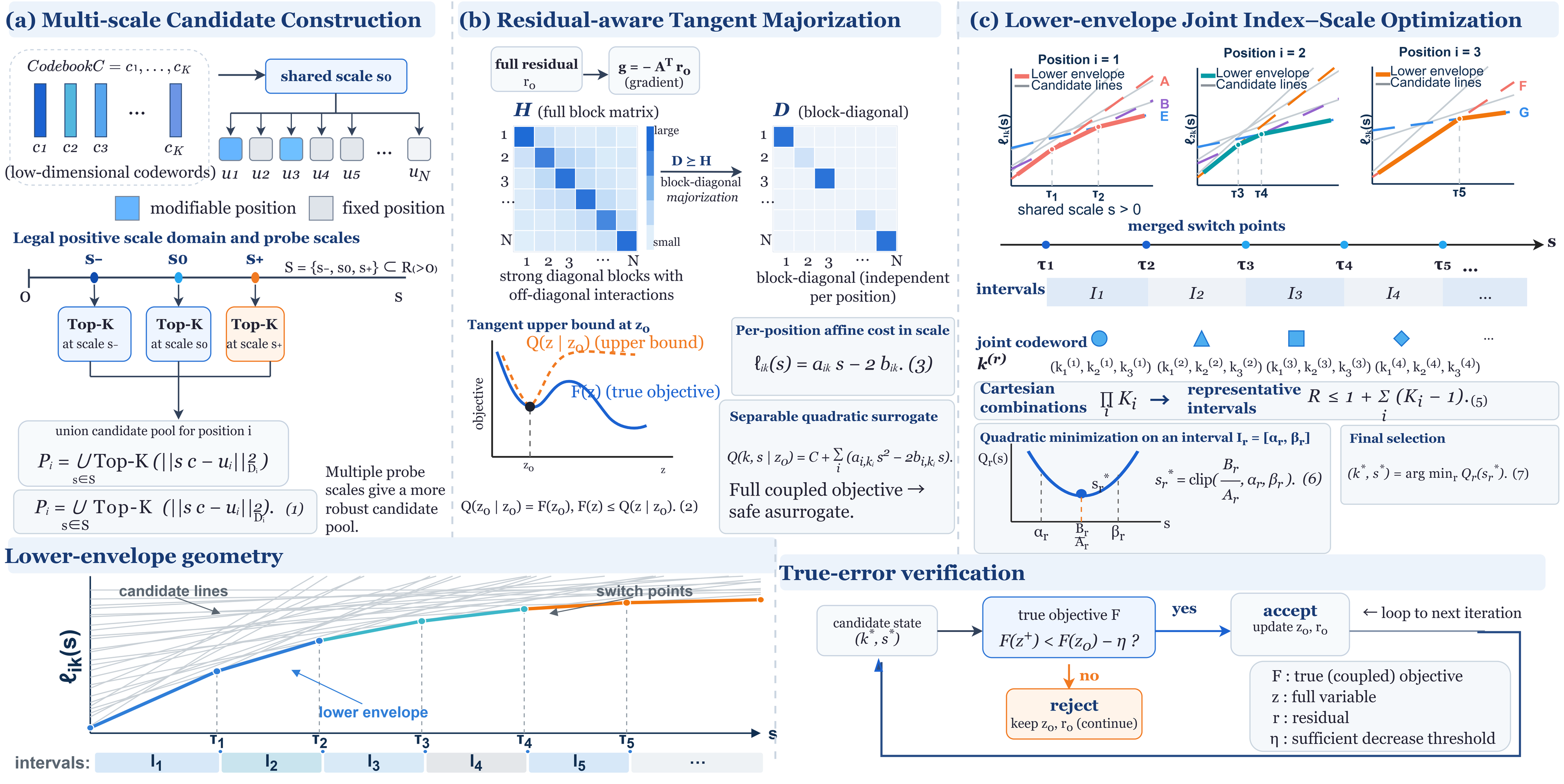}
\caption{
Overview of \method{}. (a) Multi-scale probes construct candidate pools.
(b) A residual-aware tangent majorizer yields separable
quadratic costs and affine comparisons over the shared scale.
(c) Lower-envelope breakpoints define intervals for joint
index--scale optimization.
The lower panels illustrate the envelope geometry and
true-error verification.
Proofs and boundary cases appear in
Appendix~\ref{app:envelope_proof}.
}
\label{fig:method-overview}
\end{figure*}

\paragraph{Overview.}
We propose \method{} to improve low-bit VQ through coordinated index selection under an existing shared scale, without increasing the base-codeword dimension or introducing new stored parameters. As illustrated in Figure~\ref{fig:method-overview}, the method alternates between two steps. First, a residual-aware quadratic majorizer controls cross-position interactions, while multiple scale probes provide codeword candidates beyond those preferred at the current scale. Second, a lower-envelope solver jointly selects hard indices and their shared positive scale by searching finitely many scale intervals. Every proposed update is checked against the true calibration layer-output error. The complete schedule and pseudocode appear in Appendix~\ref{app:alg:schedule}, and a checkable coordinated-update example is given in Appendix~\ref{app:example}.

\subsection{Residual-Aware Majorization and Candidate Construction}
\label{sec:surrogate}
For a selected scale group, let $z_0$ concatenate its current reconstructed weight vectors and let $\delta=z-z_0$. With the group-restricted output operator $\mathcal A$ and full residual $r_0$ defined in Section~\ref{sec:preliminaries}, set $H=\mathcal A^\top\mathcal A$ and $g=-\mathcal A^\top r_0$. The exact layer-output objective is
\begin{equation}
F(z_0+\delta)=F(z_0)+2g^\top\delta+\delta^\top H\delta.
\label{eq:quad_expand}
\end{equation}
The vector $g$ captures how each group-wise change interacts with the \emph{entire} current layer-output residual, including errors contributed by groups that are held fixed. In contrast, scoring each candidate solely against its original unquantized weight would ignore this residual context. Meanwhile, the off-diagonal blocks $H_{ij}$ describe how simultaneous changes at distinct positions interact under the calibration inputs. Directly optimizing all of these interactions is difficult when indices must remain discrete. We therefore retain the exact first-order term while replacing the cross-position quadratic curvature with the block-diagonal majorizer
\begin{equation}
\begin{gathered}
D_i=H_{ii}+\left(\sum_{j\ne i}\|H_{ij}\|_2+\varepsilon\right)I_d,\\
D=\operatorname{blkdiag}(D_1,\ldots,D_M),\quad\varepsilon>0.
\end{gathered}
\label{eq:D}
\end{equation}
The block Cauchy--Schwarz inequality gives $D-H\succeq\varepsilon I$, yielding the tangent separable upper bound~\cite{mmtutorial}
\begin{equation}
\begin{aligned}
Q(z\mid z_0)={}&F(z_0)+2g^\top(z-z_0)\\
&+(z-z_0)^\top D(z-z_0).
\end{aligned}
\label{eq:Q}
\end{equation}
The construction follows from bounding each off-diagonal interaction by its spectral norm, so it requires no manually chosen global curvature constant. In particular, $D_i\succ0$, and all interactions are upper-bounded without discarding the gradient induced by the full residual. Thus $F(z)\le Q(z\mid z_0)$ with equality and matching continuous-weight gradients at $z_0$; Appendix~\ref{app:blockmajorization} proves these properties and gives the exact surrogate gap. As with other majorization methods, the bound can be conservative, and the resulting minimizer is not automatically optimal for the true objective.

Let $h_i=D_i z_{0i}-g_i$. Substituting the shared-scale reconstruction $z_i=sc_{ik_i}$ into Eq.~\ref{eq:Q} yields
\begin{equation}
Q(k,s\mid z_0)=C+\sum_{i=1}^{M}
\left(a_{ik_i}s^2-2b_{ik_i}s\right),
\label{eq:Qks}
\end{equation}
where $a_{ik}=c_{ik}^\top D_i c_{ik}\ge0$, $b_{ik}=c_{ik}^\top h_i$, and $C$ is independent of $(k,s)$. To avoid limiting selection to a single scale, we build each mutable position's pool from the top-$K$ codewords under $\|sc-u_i\|_{D_i}^2$ at multiple probes $s\in\Sigma$, where $u_i=z_{0i}-D_i^{-1}g_i$ and $\|v\|_{D_i}^2=v^\top D_i v$. The probe set includes the current scale and explores other points in the admissible range, since a codeword that ranks poorly at the current scale may become preferable after coordinated scale movement. The union retains the current codeword and is capped at $K_{\max}$, prioritizing repeated probe hits when truncation is necessary; nonmutable positions retain singleton pools. All candidates still belong to the fixed original codebook: probes expand the \emph{search coverage} rather than the deployed representation or the number of stored scales. Moreover, all $M$ positions remain in Eq.~\ref{eq:Qks} even if only $m$ indices may change, because the common scale also changes their reconstructions. The full candidate-selection rule and the group-activation schedule are specified in Appendices~\ref{app:alg:candidates} and~\ref{app:alg:schedule}, respectively.

\subsection{Lower-Envelope Joint Index--Scale Optimization}
\label{sec:envelope}
Although the majorizer removes explicit cross-position
quadratic interactions, the shared scale still couples
codeword assignments across positions.
For a fixed positive scale $s$, candidate choices in
Eq.~\ref{eq:Qks} separate by position.
Dividing each candidate-dependent cost by $s>0$
preserves its ordering and yields the affine lines
\begin{equation}
\ell_{ik}(s)=a_{ik}s-2b_{ik},\qquad
L_i(s)=\min_{k\in\mathcal P_i}\ell_{ik}(s).
\label{eq:lines}
\end{equation}
Consequently, $\min_k Q(k,s\mid z_0)=C+s\sum_i L_i(s)$. Importantly, division by $s$ is used only to compare codeword candidates: the full quadratic dependence on the stored scale remains in the optimization objective. For each position, we sort its candidate lines by slope, remove dominated equal-slope lines, and maintain intersections between adjacent active lines to construct the lower envelope. We then merge the active switching points across all positions with the scale-domain boundaries. Between consecutive switches, each position's minimizing codeword stays unchanged, defining one representative \emph{joint} assignment $k^{(r)}$. This arrangement coordinates otherwise independent candidate rankings through their shared scale without enumerating every combination. If position $i$ has $K_i=|\mathcal P_i|$ candidates, the number of open intervals satisfies
\begin{equation}
R\le 1+\sum_{i=1}^{M}(K_i-1).
\label{eq:R}
\end{equation}
Thus the number of representative combinations grows at most linearly in the pool sizes, rather than as $\prod_i K_i$. For example, 20 mutable positions with eight candidates each produce no more than 141 representative intervals; singleton positions contribute no new switching points. Simultaneous switches may further reduce the number. This is a structural bound on interval enumeration, not on the cost of building the majorizer or ranking codebook entries at the probes.

For an interval closure $I_r=[\alpha_r,\beta_r]$, let $A_r=\sum_i a_{i,k_i^{(r)}}$ and $B_r=\sum_i b_{i,k_i^{(r)}}$. Its full surrogate remains quadratic, $Q_r(s)=C+A_rs^2-2B_rs$. For continuous scales and $A_r>0$, the constrained minimizer is
\begin{equation}
s_r^*=\operatorname{clip}(B_r/A_r,\alpha_r,\beta_r).
\label{eq:sr}
\end{equation}
The interval coefficients $A_r$ and $B_r$ can be updated incrementally at switching events, without recomputing the sums for every representative. For stored floating-point scales, we instead compare the closest legal values within each interval, explicitly handling switching boundaries and intervals without representable scales. At a switching boundary, neighboring representatives tie in the surrogate, so deterministic tie handling suffices for its minimum; tied combinations need not have identical \emph{true} errors. Appendix~\ref{app:envelope_proof} details legal-scale and degenerate cases. Comparing interval minima globally solves the \emph{current surrogate over the current candidate pools}, not the original coupled objective over the full codebook (Theorem~\ref{thm:pool_opt}).

The selected indices and scale are converted to their actual storage format. An update $z_+$ is accepted only if $F(z_+)\le F(z_0)-\eta$, where $\eta>0$ is the prescribed decrease threshold. Since the current state belongs to the pools, exact surrogate minimization implies
\begin{equation}
F(z_+)\le Q(z_+\mid z_0)\le F(z_0)
\label{eq:nonincrease}
\end{equation}
\emph{before} thresholded acceptance, under exact arithmetic. True-error verification protects against rounding and surrogate-solver approximations. Importantly, an exact surrogate optimum is guaranteed to be non-increasing under the true objective, but a surrogate can still be too conservative to propose a genuinely improving joint move; the method does not claim unconditional escape from every coordinate-wise stall. Accepted changes update the full residual before the next scheduled group is processed, and the search stops after a sweep without accepted changes or at the prescribed iteration limit. Appendix~\ref{app:acceptance_proof} derives the surrogate gap and the precise descent conditions, while Appendix~\ref{app:alg:complexity} distinguishes envelope complexity from the cost of curvature, probe scoring, and true-error verification.
After joint search, we optionally freeze hard indices and fine-tune the \emph{existing} scales on the same calibration data. This stage does not increase the number of stored scales and is evaluated in Table~\ref{tab:finetuning}. Per-model breakdowns appear in Appendix~\ref{app:all-results}.
% 实验
\section{Experiments}
\label{sec:experiments}

\subsection{Experimental Setup}

\paragraph{Models and datasets.}
We evaluate four Qwen3~\cite{qwen3} models (4B, 8B, 14B, and 32B)
and two Llama-3.1~\cite{llama3} models (8B-Instruct and 70B-Instruct).
Qwen3.8-27B~\cite{qwen38} is evaluated separately with thinking enabled.
Perplexity is measured on WikiText-2~\cite{wikitext} and
C4~\cite{c4} with sequence length 2,048.
We use LM Evaluation Harness v0.4.4~\cite{lmeval} for zero-shot
ARC-Challenge and ARC-Easy~\cite{arc},
HellaSwag~\cite{hellaswag}, LAMBADA OpenAI~\cite{lambada},
PIQA~\cite{piqa}, BoolQ~\cite{boolq}, MMLU~\cite{mmlu},
and WinoGrande~\cite{winogrande}.
IFEval~\cite{ifeval} is used only for instruction/chat models
with an official chat template.

\paragraph{Quantization protocol.}
The main setting uses $d=6$, $K=4096$, and a 12-bit index
(2 bits per weight), with one FP16 group scale per 128
input columns ($\approx 0.125$ bit per weight).
Each round selects at most four existing scale groups per row.
We calibrate each model using 320 RedPajama-Data-1T~\cite{redpajama}
sequences of length 4,096.
After joint search, we freeze hard indices and fine-tune
existing scales on the same calibration data.
Detailed results for Table~\ref{tab:main-ppl} appear in
Appendix~\ref{app:all-results}.
Table~\ref{tab:finetuning} evaluates scale fine-tuning,
while Table~\ref{tab:ablation} compares scale and
assignment optimization from identical initializations.

\paragraph{Baselines and hardware.}
We compare against BF16, PV-Tuning~\cite{pvtuning},
QuIP\#~\cite{quipsharp}, QTIP~\cite{qtip},
NoWag~\cite{nowag}, AQLM~\cite{aqlm}, and
LLVQ~\cite{llvq}.
All quality evaluations and joint-search experiments
are performed on NVIDIA A100-SXM4-80GB GPUs.
Decode throughput in Table~\ref{tab:efficiency}
is measured on an NVIDIA GeForce RTX 4090.

\begin{table*}[t]
\caption{
Main results on six models. PPL.AVG is the arithmetic mean of WikiText-2 and C4 perplexities. ACC.AVG is the macro-average of ARC-Challenge, ARC-Easy, HellaSwag, LAMBADA OpenAI, PIQA, and WinoGrande. Bold denotes the best result among quantized methods.
}
\label{tab:main-ppl}
\centering
\small
\setlength{\tabcolsep}{3pt}
% Two stacked panels preserve the six-model table at a readable size.
\begin{tabular*}{0.85\linewidth}{@{\extracolsep{\fill}}l*{3}{cc}}
\toprule
& \multicolumn{2}{c}{Qwen3-4B} & \multicolumn{2}{c}{Qwen3-8B} & \multicolumn{2}{c}{Qwen3-14B} \\
\cmidrule(lr){2-3}\cmidrule(lr){4-5}\cmidrule(lr){6-7}
Method & PPL.AVG$\downarrow$ & ACC.AVG$\uparrow$ & PPL.AVG$\downarrow$ & ACC.AVG$\uparrow$ & PPL.AVG$\downarrow$ & ACC.AVG$\uparrow$ \\
\midrule
BF16 & 15.77 & 66.79 & 12.01 & 70.37 & 10.75 & 73.77 \\
\midrule
PV-Tuning & 17.10 & 59.57 & 12.90 & 66.84 & 11.36 & 70.92 \\
QuIP\# & 21.48 & 57.03 & 15.76 & 65.32 & 13.37 & 69.65 \\
QTIP & 18.20 & 60.30 & 13.91 & 66.80 & 12.09 & \textbf{71.50} \\
NoWag & 23.46 & 57.19 & 15.96 & 65.73 & 13.32 & 69.63 \\
AQLM & 19.25 & 60.08 & 14.28 & 67.12 & 12.39 & 71.03 \\
LLVQ & 16.13 & 60.77 & \textbf{11.57} & 67.35 & 11.32 & 71.28 \\
\method{} & \textbf{15.83} & \textbf{61.42} & 12.20 & \textbf{68.61} & \textbf{10.80} & 71.01 \\
\midrule
\addlinespace
& \multicolumn{2}{c}{Qwen3-32B} & \multicolumn{2}{c}{Llama-3.1-8B-Inst.} & \multicolumn{2}{c}{Llama-3.1-70B-Inst.} \\
\cmidrule(lr){2-3}\cmidrule(lr){4-5}\cmidrule(lr){6-7}
Method & PPL.AVG$\downarrow$ & ACC.AVG$\uparrow$ & PPL.AVG$\downarrow$ & ACC.AVG$\uparrow$ & PPL.AVG$\downarrow$ & ACC.AVG$\uparrow$ \\
\midrule
BF16 & 9.56 & 74.81 & 8.93 & 73.55 & 5.89 & 78.37 \\
\midrule
PV-Tuning & 10.18 & 74.51 & 10.78 & 64.90 & 8.33 & 73.54 \\
QuIP\# & 12.15 & 73.42 & 12.30 & 64.71 & 8.39 & 71.45 \\
QTIP & 10.51 & 74.21 & 11.10 & \textbf{69.52} & 7.78 & 74.01 \\
NoWag & 11.27 & 74.29 & 12.89 & 67.67 & 8.68 & 73.09 \\
AQLM & 10.68 & 74.63 & 12.25 & 55.55 & 8.19 & 69.01 \\
LLVQ & 10.01 & \textbf{74.70} & 10.86 & 68.75 & 7.77 & 73.47 \\
\method{} & \textbf{9.87} & 74.67 & \textbf{10.56} & 69.16 & \textbf{7.73} & \textbf{74.05} \\
\bottomrule
\end{tabular*}
\vspace{-0.4cm}
\end{table*}

\subsection{Main Results}

\begin{table}[t]
\centering
\caption{
Same-start ablation of shared-scale optimization (Scale)
and hard-index assignment (Assign).
Avg. denotes the mean accuracy across eight zero-shot tasks.
}
\label{tab:ablation}
\small
\setlength{\tabcolsep}{5pt}
\begin{tabular}{ccccc}
\toprule
Model & Scale & Assign & Wiki2$\downarrow$ & Avg.$\uparrow$ \\
\midrule
\multirow{3}{*}{Qwen3-4B}
& $\times$ & $\times$ & 22.69 & 58.81 \\
& $\checkmark$ & $\times$ & 21.01 & 59.32 \\
& $\checkmark$ & $\checkmark$ & \textbf{13.68} & \textbf{62.58} \\
\midrule
\multirow{3}{*}{Qwen3-8B}
& $\times$ & $\times$ & 14.41 & 66.44 \\
& $\checkmark$ & $\times$ & 13.85 & 67.19 \\
& $\checkmark$ & $\checkmark$ & \textbf{9.76} & \textbf{69.43} \\
\bottomrule
\end{tabular}
\vspace{-0.3cm}
\end{table}

Table~\ref{tab:main-ppl} compares \method{} with
quantization baselines at a 2-bit target across six LLMs. \method{} achieves the lowest average perplexity on five of
six models and the highest average downstream accuracy on three,
demonstrating competitive performance across model families.
On Qwen3-4B, it improves PPL.AVG from 16.13 to 15.83 and
ACC.AVG from 60.77 to 61.42 compared with LLVQ.
Although LLVQ and QTIP retain advantages on certain models
or metrics, \method{} delivers a competitive
balance between language modeling and downstream accuracy.
These results highlight the effectiveness of coordinated
selection with low-dimensional codewords.
Detailed task-wise results are provided in
Appendix~\ref{app:all-results}, while storage accounting,
including indices, codebooks, and scales, is discussed in
Appendix~\ref{app:impl}.

\begin{table}[t]
\centering
\caption{
Qwen3.8-27B with thinking enabled.
Bold denotes the best quantized result.
}
\label{tab:qwen38}
\small
\setlength{\tabcolsep}{3pt}
\resizebox{\columnwidth}{!}{%
\begin{tabular}{cccccc}
\toprule
Method & Wiki2$\downarrow$ & C4$\downarrow$ &
AIME25$\uparrow$ & LCBv6$\uparrow$ & GPQA-D$\uparrow$ \\
\midrule
BF16 & 7.05 & 11.45 & 100.00 & 85.71 & 89.90 \\
QTIP & 8.29 & 13.76 & 93.33 & 73.65 & 81.11 \\
LLVQ & 8.22 & 13.85 & 93.33 & 72.28 & 80.48 \\
\method{} & \textbf{7.84} & \textbf{13.33} &
\textbf{93.33} & \textbf{74.92} & \textbf{82.83} \\
\bottomrule
\end{tabular}%
}
\vspace{-0.3cm}
\end{table}

\subsection{Ablation Studies}

\paragraph{Scale and assignment.}

Table~\ref{tab:ablation} evaluates shared-scale refinement
and hard-index assignment from identical initializations.
Avg. denotes the mean accuracy across eight
zero-shot tasks: ARC-Challenge, ARC-Easy, HellaSwag,
LAMBADA OpenAI, PIQA, BoolQ, MMLU, and WinoGrande.
Scale-only optimization reduces WikiText-2 perplexity
from 22.69/14.41 to 21.01/13.85 on Qwen3-4B/8B.
In contrast, joint optimization reduces perplexity
to 13.68/9.76 and improves average zero-shot accuracy
from 59.32/67.19 to 62.58/69.43.
These gains demonstrate that coordinated hard-index updates,
rather than scale refinement alone, account for most of
the improvement by enabling joint changes across positions.

% 表 3 在开启思考模式的 Qwen3.8-27B 上比较推理和代码能力。
\paragraph{Thinking-mode evaluation.} 
Table~\ref{tab:qwen38} evaluates Qwen3.8-27B with thinking enabled. \method{} improves WikiText-2/C4 perplexity, LiveCodeBench v6, and GPQA-D over both quantized baselines~\cite{livecodebench,gpqa}, and it matches them on AIME 2025~\cite{aime2025}. Relative to QTIP, WikiText-2/C4 fall from 8.29/13.76 to 7.84/13.33, LiveCodeBench v6 rises from 73.65 to 74.92, and GPQA-D rises from 81.11 to 82.83; AIME 2025 is 93.33 for all three quantized methods.

\subsection{Efficiency}
\paragraph{Quantization time and inference throughput.}
Table~\ref{tab:efficiency} reports offline quantization time
on an A100-SXM4-80GB GPU and inference throughput on an RTX 4090.
Quantization takes 1.3 hours for Qwen3-4B and 3.5 hours for
Qwen3-32B.
On Llama-3.1-8B-Instruct, prefill throughput remains comparable
to BF16 (2230.2 versus 2217.6 tok/s), while decode throughput
increases from 21.9 to 36.7 tok/s, achieving a $1.68\times$
speedup at a 2K-token context.
These gains benefit from low-dimensional codebook lookup,
while joint optimization is performed entirely offline.
\begin{table}[t]
\centering
\caption{
Offline quantization time on an A100-SXM4-80GB GPU
and Llama-3.1-8B-Instruct inference throughput on
an RTX 4090 with a 2K-token context.
}
\label{tab:efficiency}
\small
\setlength{\tabcolsep}{3pt}
\begin{tabular*}{\columnwidth}{@{\extracolsep{\fill}}ccccc}
\toprule
\multicolumn{5}{c}{\textbf{Offline Quantization Time}} \\
\midrule
Model & Q3-4B & Q3-8B & Q3-14B & Q3-32B \\
Time (h)$\downarrow$ & 1.3 & 1.6 & 2.0 & 3.5 \\
\midrule
\multicolumn{5}{c}{\textbf{Inference Throughput}} \\
\midrule
Method & \multicolumn{2}{c}{Prefill (tok/s)$\uparrow$}
       & \multicolumn{2}{c}{Decode (tok/s)$\uparrow$} \\
BF16 & \multicolumn{2}{c}{2217.6}
     & \multicolumn{2}{c}{21.9} \\
\method{} & \multicolumn{2}{c}{\textbf{2230.2}}
          & \multicolumn{2}{c}{\textbf{36.7}} \\
\bottomrule
\end{tabular*}
\vspace{-0.4cm}
\end{table}

\paragraph{Base-codeword dimension.}
Table~\ref{tab:dimension} examines the quality--rate trade-off
across different base-codeword dimensions.
Increasing $d$ to 8 improves all evaluated metrics but
raises the bitrate to 2.59 bpw.
Under an approximately 2-bpw budget, $d=4$ and $d=5$
achieve lower WikiText-2 perplexity, whereas $d=6$
consistently outperforms both on all four downstream tasks.
We therefore adopt $d=6$ as the default setting,
balancing downstream accuracy and storage efficiency.
\begin{table}[t]
\centering
\caption{
Effect of base-codeword dimension $d$ on Qwen3-4B.
The default $d=6$ balances accuracy and bitrate
under an approximately 2-bpw budget.
}
\label{tab:dimension}
\small
\setlength{\tabcolsep}{2.5pt}
\begin{tabular*}{\columnwidth}{@{\extracolsep{\fill}}ccccccc}
\toprule
$d$ & bpw & Wiki2$\downarrow$ & ARC-C$\uparrow$ &
ARC-E$\uparrow$ & HellaSwag$\uparrow$ & MMLU$\uparrow$ \\
\midrule
4 & 2.01 & 12.96 & 41.27 & 68.90 & 62.11 & 53.23 \\
5 & 2.02 & 12.74 & 42.50 & 69.46 & 62.78 & 53.89 \\
6 & 2.04 & 13.68 & 42.92 & 70.12 & 63.16 & 55.01 \\
8 & 2.59 & 11.87 & 43.27 & 70.73 & 64.28 & 55.71 \\
\bottomrule
\end{tabular*}
\vspace{-0.3cm}
\end{table}

\paragraph{End-to-end scale fine-tuning.}
Table~\ref{tab:finetuning} evaluates scale fine-tuning
after joint search with hard indices frozen.
Fine-tuning consistently reduces WikiText-2 perplexity
across all four Qwen3 models, particularly on Qwen3-14B
(9.78 to 8.51), and improves ARC-C accuracy.
HellaSwag and WinoGrande show overall gains,
with minor regressions on certain models.
Unlike the coordinated index updates in
Table~\ref{tab:ablation}, this stage refines only
existing scales, providing complementary improvements
rather than replacing joint selection.
This ablation is distinct from the endpoints
in Table~\ref{tab:main-ppl}.
Additional task-wise results appear in
Appendix~\ref{app:all-results}.
\begin{table}[t]
\centering
\caption{
Ablation of end-to-end scale fine-tuning on Qwen3.
Each entry reports \emph{fine-tuned / not fine-tuned}.
}
\label{tab:finetuning}
\small
\setlength{\tabcolsep}{2pt}
\resizebox{\columnwidth}{!}{%
\begin{tabular}{ccccc}
\toprule
Model & Wiki2$\downarrow$ & ARC-C$\uparrow$ &
HellaS$\uparrow$ & WinoG$\uparrow$ \\
\midrule
Q3-4B  & 12.18/12.69 & 42.92/42.63 &
63.17/62.28 & 62.90/62.39 \\
Q3-8B  & 9.76/10.17 & 54.52/53.87 &
69.80/69.64 & 68.67/68.80 \\
Q3-14B & 8.51/9.78 & 53.24/52.76 &
75.30/75.47 & 71.98/71.01 \\
Q3-32B & 7.74/8.06 & 60.32/59.78 &
78.17/78.11 & 76.09/76.30 \\
\bottomrule
\end{tabular}%
}
\vspace{-0.4cm}
\end{table}

\section{Conclusion}
We presented \method{}, a low-bit VQ framework preserving
independently accessible, locally rewritable low-dimensional
codewords through coordinated shared-scale selection.
To address coordinate-wise stalls after scale re-optimization,
\method{} combines multi-scale candidate pools, a residual-aware
tangent separable upper bound, and lower-envelope search
to jointly optimize hard indices and scales
without exhaustive enumeration.
Same-start ablations show improvements
over scale-only refinement on Qwen3-4B/8B.
At approximately 2\,bpw, \method{} outperforms QTIP and
LLVQ on language modeling, coding, and selected reasoning
benchmarks for Qwen3.8-27B.
These findings establish joint selection as a promising
alternative to increasing codeword dimensionality.

\section*{Limitations}
\label{sec:limitations}

CoS-VQ optimizes hard indices and shared scales within
finite candidate pools, so its theoretical optimality
guarantee applies to the surrogate objective rather than
the original layer-output error. Although true-error
verification ensures non-increasing accepted updates,
the surrogate may be too conservative to escape certain
coordinate-wise stalls. Moreover, candidate quality depends
on the multi-scale probe coverage and the chosen codebook.
Our experiments focus on selected LLM families and
benchmarks, leaving broader architectures and extremely
low-bit configurations for future evaluation.
Finally, inference efficiency depends on hardware and
codebook lookup implementations, and performance across
different deployment environments remains to be explored.

\section*{Ethical Considerations}
\label{sec:ethics}

CoS-VQ reduces the memory requirements of existing LLMs
through post-training quantization without modifying their
underlying training objectives. Improved compression may
facilitate more accessible and resource-efficient deployment,
although actual energy savings depend on hardware and
serving conditions and have not been independently measured.
Quantization can alter model outputs, potentially affecting
factual reliability, fairness, and safety-related behavior.
Our evaluations do not constitute dedicated safety, bias,
or privacy assessments. Furthermore, reduced deployment
costs may also facilitate harmful applications of LLMs.
We therefore recommend retaining appropriate access controls
and safety safeguards, together with task-specific evaluations,
before practical deployment.

\bibliography{sec/references}

\clearpage

\appendix
\section{Appendix}
\label{sec:appendix}
\subsection{Joint Search Procedure}
\label{app:algorithm}

\subsubsection{Scheduling and Full-Group Scope}
\label{app:alg:schedule}
The joint-search unit is a \emph{complete existing scale group}, not just the subset of mutable indices. Let $\mathcal G_h$ contain $M_h$ positions sharing one stored positive scale $s_h$. During one update, a scheduled subset $\mathcal J_h\subseteq\mathcal G_h$ of at most $m$ positions may change their hard indices. For $i\notin\mathcal J_h$, the candidate pool is the singleton $\{k_i^{(0)}\}$. Nevertheless, all $M_h$ positions contribute to the coefficients in Eq.~\ref{eq:Qks} and to true-error verification, because a change in $s_h$ modifies the reconstruction of even a fixed-index position. Group and mutable-position schedules are fixed independently of the interval optimizer; the surrogate-optimality claim is conditional on the pools they select.

To implement the search budget \texttt{groups\_per\_row=4}, we partition eligible groups into an ordered sequence of batches $\mathfrak B=(\mathcal B_1,\ldots,\mathcal B_J)$ such that each batch activates at most four \emph{existing} scale groups per matrix row. A complete sweep visits all batches in order. If more than four groups are eligible in a row, later batches process the remaining groups; none is silently removed. The parameter limits simultaneous search scope and does not create extra stored scales. After each accepted update, the current residual is updated before statistics are formed for the next group.

\subsubsection{Multi-Scale Candidate-Pool Construction}
\label{app:alg:candidates}
For a selected group, let $z_{0i}$ be the current reconstruction at position $i$ and let $g_i,D_i$ be the corresponding residual-gradient and majorizer blocks. As $D_i\succ0$, the completed-square form of the position-wise surrogate is
\begin{equation}
\begin{aligned}
&2g_i^\top(z_i-z_{0i})+(z_i-z_{0i})^\top D_i(z_i-z_{0i})\\
&\qquad=\|z_i-u_i\|_{D_i}^{2}-g_i^\top D_i^{-1}g_i,\\
&u_i=z_{0i}-D_i^{-1}g_i,\qquad \|v\|_{D_i}^{2}=v^\top D_i v.
\end{aligned}
\label{eq:surrogate_center}
\end{equation}
The additive term is independent of the codeword and scale. For each $i\in\mathcal J_h$, construct a finite probe set $\Sigma$ containing the current scale $s_h$, the two domain endpoints, and a predetermined number of geometrically spaced interior probes (duplicates removed). The probes lie in the positive scale domain but need not themselves be stored as candidate final scales. At each probe $\sigma\in\Sigma$, select the top $K$ original codewords by
\begin{equation}
\operatorname{score}_{ik}(\sigma)
=\|\sigma c_{ik}-u_i\|_{D_i}^{2}.
\label{eq:probe_score}
\end{equation}
Take the union of probe shortlists and explicitly insert the current index $k_i^{(0)}$. If the union exceeds $K_{\max}$, retain $k_i^{(0)}$ and select the remaining candidates by decreasing probe-occurrence count, breaking ties by their best probe rank and then the fixed codeword index. We assume $K_{\max}\ge1$ and use $K_i=|\mathcal P_i|\le K_{\max}$. Positions outside $\mathcal J_h$ have $K_i=1$. This rule defines a deterministic truncated pool; its role is to avoid restricting all candidate rankings to $s_h$. It does not guarantee that the pool contains the true or surrogate optimum over the \emph{entire} original codebook.

\subsubsection{Full Joint-Search Pseudocode}
\label{app:alg:pseudocode}
Algorithm~\ref{alg:joint} spells out the alternating procedure. The algorithmic state $\theta=(k,s)$ stores the hard indices and existing group scales, while $z_h(\theta)$ denotes the concatenated \emph{reconstructed weight vectors} in group $h$. The operator $\mathcal A_h$ maps a reconstruction change in group $h$ to its vectorized layer-output change. Therefore the updated residual is exactly $r^+=r-\mathcal A_h(z_h^+-z_{0h})$ while the other groups are fixed. The packing step precedes true-error evaluation so that the verified candidate matches the deployed storage format.

\begin{algorithm*}[!tbp]
\small
\caption{\method{}: coordinated hard-index and shared-scale search}
\label{alg:joint}
\begin{algorithmic}[1]
\Require $W,X$, stored state $\theta^{(0)}$, fixed codebooks $\{\mathcal C_i\}$, ordered group batches $\mathfrak B$, mutable-position schedule $\{\mathcal J_h\}$, probe rule $\Sigma$, $K,K_{\max}$, legal scale sets $\{\mathcal S_h\}$, sweep limit $T$, threshold $\eta>0$
\Ensure Updated stored state $\theta=(k,s)$
\State $\theta\gets\theta^{(0)}$; $r\gets\operatorname{vec}(WX-\widehat W(\theta)X)$
\For{$t=1,\ldots,T$}
    \State $\mathtt{accepted}\gets\textbf{false}$
    \For{batch $\mathcal B$ in the fixed sequence $\mathfrak B$}
        \For{group $h\in\mathcal B$ in the fixed order}
            \State $z_{0h}\gets z_h(\theta)$; $\mathcal A_h\gets\operatorname{GroupOperator}(X,\mathcal G_h)$
            \State $g_h\gets-\mathcal A_h^\top r$; $H_h\gets\mathcal A_h^\top\mathcal A_h$
            \State $D_h\gets\operatorname{BlockMajorizer}(H_h)$ by Eq.~\ref{eq:D}
            \State $\{\mathcal P_i\}_{i\in\mathcal G_h}\gets\operatorname{ProbeUnion}(z_{0h},g_h,D_h,\Sigma,K,K_{\max},\mathcal J_h)$
            \State $(k_h^+,s_h^+)\gets\operatorname{EnvelopeSearch}(z_{0h},g_h,D_h,\{\mathcal P_i\},\mathcal S_h)$
            \State $\theta_h^+\gets\operatorname{Pack}_{\mathrm{stored}}(k_h^+,s_h^+)$; $z_h^+\gets z_h(\theta_h^+)$
            \State $r^+\gets r-\mathcal A_h(z_h^+-z_{0h})$
            \If{$\|r^+\|_2^2\le\|r\|_2^2-\eta$}
                \State $\theta|_h\gets\theta_h^+$; $r\gets r^+$; $\mathtt{accepted}\gets\textbf{true}$
            \EndIf
        \EndFor
    \EndFor
    \If{\textbf{not} $\mathtt{accepted}$} \State \textbf{break} \EndIf
\EndFor
\State \Return $\theta$
\end{algorithmic}
\end{algorithm*}

\paragraph{Legal scale solving.}
The interval solver constructs the exact lower envelopes of Eq.~\ref{eq:lines}, intersects their switching points with $\mathcal S_h$, and evaluates each interval's constrained quadratic minimizer. For floating-point storage, $\mathcal S_h$ denotes the \emph{finite representable} positive scales in the permitted range. Each interval is evaluated using its legal predecessor/successor nearest to the continuous vertex, together with any legal endpoints. When a breakpoint is exactly representable, either adjacent optimal envelope representative attains the same surrogate value there; deterministic tie handling selects one. An interval containing no legal scale is skipped. If only one legal scale exists, codeword comparisons are performed directly at that scale. Full details and a proof appear in Appendix~\ref{app:envelope_proof}.

\paragraph{Stopping and acceptance.}
The check against $\|r\|_2^2$ is mandatory, not a substitute for surrogate minimization. In exact arithmetic with exact upper-bound solving, a chosen candidate cannot increase the true error, but it need not meet the strict decrease threshold $\eta$. Rejected candidates leave both $\theta$ and $r$ unchanged. A sweep terminates only after all scheduled batches have been considered. For a fixed absolute $\eta>0$, the nonnegative objective bounds the number of accepted moves by $\lfloor F(\theta^{(0)})/\eta\rfloor$ (up to the convention at equality); the explicit sweep limit $T$ also bounds runtime. If using a relative threshold in practice, its numerical definition must be recorded and the absolute bound applies only when a positive fixed absolute component is retained.

\subsubsection{Complexity and Implementation Notes}
\label{app:alg:complexity}
Let $K_i=|\mathcal P_i|$ and $B$ be the number of distinct interior switching points in the legal scale range, with $B\le\sum_i(K_i-1)$. Once $g,H,D$ and the coefficient pairs $(a_{ik},b_{ik})$ are available, sorting lines by slope and building every position's lower envelope takes
\begin{equation}
O\!\left(\sum_i K_i\log K_i\right),
\label{eq:envelope_cost}
\end{equation}
while merging the switching points by a global sort takes $O(B\log B)$. Initializing $A_r,B_r$ takes $O(M)$; updating them across adjacent intervals takes $O(B)$, since only positions switching at that event change their contributions. Constrained continuous-scale minimization then takes $O(B+M)$ beyond the breakpoint sort. Recovering the best combination requires storing an initial representative and replaying the switching events; a full index vector for every interval is unnecessary. Discrete-scale predecessor/successor queries add their actual lookup cost (constant time with direct representable-float neighbors, or logarithmic time in the size of a sorted legal-scale array).

These bounds describe \emph{the envelope stage}, not the total algorithm. Full runtime must additionally account for forming or applying the group operator, evaluating the residual and $H=\mathcal A^\top\mathcal A$, computing the spectral block norms in Eq.~\ref{eq:D}, solving for probe centres, scoring the original codebook at all probe scales, generating coefficients, and verifying the packed candidate on the true objective. In a direct dense implementation, evaluating $c^\top D_i c$ costs $O(d^2)$ per codeword after $D_i$ is available; efficient implementations may exploit operator or block structure. Likewise, explicitly storing all cross-position blocks of $H$ can dominate memory when $M$ is large. Neither this overhead nor optional post-search scale fine-tuning is included in the envelope-only complexity. The base method verifies only the single best surrogate candidate per group; evaluating all interval representatives with the true objective would be a distinct algorithmic variant.
\subsection{A Checkable Coordinated-Update Instance}
\label{app:example}

Consider two scalar positions with targets $w_1=w_2=1$. The current hard-codeword assignment is $(a,b)=(1,6/5)$; the available codewords are $\mathcal C_1=\{1,3\}$ and $\mathcal C_2=\{6/5,3\}$. Both positions use the same scale $s\in[1/10,2]$. For each fixed pair $(a,b)$, the reconstruction error is
\begin{equation}
E(a,b;s)=(1-sa)^2+(1-sb)^2.
\label{eq:example_obj}
\end{equation}
Differentiating yields $\partial_s E=2(a^2+b^2)s-2(a+b)$. As $a^2+b^2>0$, the unconstrained optimum and the minimized error are
\begin{equation}
\begin{gathered}
s^*(a,b)=\frac{a+b}{a^2+b^2},\\
E^*(a,b)=\frac{(a-b)^2}{a^2+b^2}.
\end{gathered}
\label{eq:example_closed}
\end{equation}
All four optimal scales lie inside the prescribed domain, so no clipping is needed. Table~\ref{tab:appendix_example} gives the exact rational values and their decimal equivalents.

\begin{table*}[h]
\centering
\small
\begin{tabular}{lccc}
\toprule
% 更新 & 组合 $(a,b)$ & 最优尺度 & 最优误差
Update & Combination $(a,b)$ & Optimal scale & Optimal error \\
\midrule
% 当前状态
Current state & $(1,6/5)$ & $55/61$ & $1/61\approx0.016393$ \\
% 只改位置 1
Change position 1 only & $(3,6/5)$ & $35/87$ & $9/29\approx0.310345$ \\
% 只改位置 2
Change position 2 only & $(1,3)$ & $2/5$ & $2/5=0.4$ \\
% 同时修改
Change both jointly & $(3,3)$ & $1/3$ & $0$ \\
\bottomrule
\end{tabular}
\caption{
% 附录 A 的可核算协同更新实例。两个单位置更新即使各自重新优化尺度也更差，而联合修改可以消除误差。
A checkable coordinated-update instance for Appendix~\ref{app:example}.
Both one-at-a-time updates remain worse even after each refits the scale, whereas a joint modification can eliminate the error.
}
\label{tab:appendix_example}
\end{table*}

At the current assignment, the best error is $1/61$. Changing only position 1 yields $9/29$, and changing only position 2 yields $2/5$: both are strictly worse \emph{even after optimally re-fitting the shared scale}. However, changing both indices to $(3,3)$ and setting $s=1/3$ gives zero error. Thus coordinate-wise stagnation need not imply joint optimality. The example isolates the scale-coupling mechanism under a Euclidean loss; the full layer-output loss additionally includes cross-position interactions. It is not a guarantee that a particular majorizer or truncated candidate pool will recover every jointly improving move. Embedding each scalar in the first coordinate of a six-dimensional vector and setting the other coordinates to zero reproduces this construction at the default base-codeword dimension $d=6$.

\subsection{Proofs and Theoretical Properties}
\label{app:theory}

\subsubsection{Exact Group Objective and Block Majorization}
\label{app:blockmajorization}
\paragraph{Restricted residual model.}
Fix all quantized positions outside the current scale group and let $z_0=(z_{01},\ldots,z_{0M})$ collect the current reconstructions of its $M$ positions. Let $\mathcal A\delta$ be the vectorized layer-output change induced by scattering the group-wise modification $\delta$ back into the quantized weight matrix. Since matrix multiplication by the fixed calibration input $X$ is linear, $\mathcal A$ is a linear operator. With $r_0=\operatorname{vec}(WX-\widehat W_0X)$,
\begin{equation}
\begin{aligned}
F(z_0+\delta)
&=\|r_0-\mathcal A\delta\|_2^2\\
&=F(z_0)-2r_0^\top\mathcal A\delta
 +\delta^\top\mathcal A^\top\mathcal A\delta.
\end{aligned}
\label{eq:app_residual_expansion}
\end{equation}
This proves Eq.~\ref{eq:quad_expand} with $g=-\mathcal A^\top r_0$ and $H=\mathcal A^\top\mathcal A\succeq0$. Crucially, the $r_0$ used in $g$ is the full current layer residual, not a residual formed only from the current scale group's original weights. Therefore, interactions with all fixed out-of-group reconstruction errors enter the first-order term, even though only the in-group variables are optimized.

\begin{proposition}[Valid upper bound and tangency]
\label{prop:majorizer}
Let $H=H^\top\succeq0$ be partitioned into position-wise $d\times d$ blocks, and define $D$ as in Eq.~\ref{eq:D} with $\varepsilon>0$. Then $D-H\succeq\varepsilon I$ and, for all group reconstruction vectors $z$,
\begin{equation}
\begin{aligned}
F(z)&\le Q(z\mid z_0),\\
Q(z_0\mid z_0)&=F(z_0),\\
\nabla_z Q(z_0\mid z_0)&=\nabla_z F(z_0).
\end{aligned}
\end{equation}
The gradient equality concerns the continuous reconstruction vectors, not differentiation through hard codeword indices.
\end{proposition}

\paragraph{Proof.}
For any $\delta=(\delta_1,\ldots,\delta_M)$, symmetry of $H$ gives
\begin{equation}
\begin{aligned}
\delta^\top H\delta
&=\sum_i\delta_i^\top H_{ii}\delta_i
 +2\sum_{i<j}\delta_i^\top H_{ij}\delta_j\\
&\le\sum_i\delta_i^\top H_{ii}\delta_i\\
&\quad+\sum_{i<j}\|H_{ij}\|_2
 (\|\delta_i\|_2^2+\|\delta_j\|_2^2)\\
&=\delta^\top(D-\varepsilon I)\delta.
\end{aligned}
\label{eq:app_block_bound}
\end{equation}
Here we used $2u^\top H_{ij}v\le 2\|H_{ij}\|_2\|u\|_2\|v\|_2\le\|H_{ij}\|_2(\|u\|_2^2+\|v\|_2^2)$. Thus $D-H\succeq\varepsilon I$. Subtracting the exact objective from Eq.~\ref{eq:Q} yields
\begin{equation}
\begin{aligned}
Q(z\mid z_0)-F(z)
&=(z-z_0)^\top(D-H)(z-z_0)\\
&\ge\varepsilon\|z-z_0\|_2^2\ge0.
\end{aligned}
\label{eq:gap}
\end{equation}
At $z=z_0$, this difference and its first derivative vanish, which establishes both tangency statements. \hfill$\square$

\paragraph{Separable coefficients.}
Expanding Eq.~\ref{eq:Q} with $\delta=z-z_0$ and using block diagonality gives
\begin{equation}
\begin{aligned}
Q(z\mid z_0)
&=C+\sum_i\bigl(z_i^\top D_i z_i-2h_i^\top z_i\bigr),\\
h_i&=D_i z_{0i}-g_i,\\
C&=F(z_0)-2g^\top z_0+z_0^\top D z_0.
\end{aligned}
\label{eq:app_coefficients}
\end{equation}
Substituting $z_i=sc_{ik_i}$ proves Eq.~\ref{eq:Qks} with $a_{ik}=c_{ik}^\top D_i c_{ik}\ge0$ and $b_{ik}=c_{ik}^\top h_i$. The codebook need not be orthogonal, and different positions may access different subsets of a shared codebook. The only required conditions are the fixed-codeword representation and a common positive scale within the selected group. Because $D_i\succeq\varepsilon I$, the surrogate centre $u_i=D_i^{-1}h_i=z_{0i}-D_i^{-1}g_i$ used in Appendix~\ref{app:alg:candidates} is well-defined.

\subsubsection{Lower Envelopes and Candidate-Pool Optimality}
\label{app:envelope_proof}
\paragraph{Reduction to affine comparisons.}
For fixed $s>0$ and fixed candidate pools, Eq.~\ref{eq:Qks} is separable in the indices. Hence
\begin{equation}
\min_{k\in\prod_i\mathcal P_i}Q(k,s\mid z_0)
=C+s\sum_i L_i(s).
\label{eq:app_min_envelope}
\end{equation}
where $L_i(s)=\min_{k\in\mathcal P_i}\ell_{ik}(s)$ and $\ell_{ik}(s)=a_{ik}s-2b_{ik}$. Dividing by $s>0$ is used \emph{only to compare candidates at a fixed scale}; the factor $s$ remains in the global scale objective. In particular, neither the scale variable nor the quadratic term is discarded from the optimization.

\paragraph{Geometry and interval bound.}
Every $\ell_{ik}$ is affine. For a given candidate $k$, the set of scales on which it is no greater than every competing line is an intersection of one-dimensional half-lines, hence an interval, a singleton, or the empty set. Sort the lines by slope, remove dominated equal-slope lines and duplicate lines using a deterministic index tie rule, and build the lower envelope using intersections of adjacent active lines. Each distinct line can occupy at most one nonempty open segment. Therefore, at position $i$, at most $K_i-1$ interior switching points remain. Pooling the distinct breakpoints across all $M$ positions gives at most $B=\sum_i(K_i-1)$ switches and $R\le B+1$ open intervals, as in Eq.~\ref{eq:R}. Simultaneous switches only decrease $R$. For example, 20 mutable positions with eight candidates each, and all other positions singletons, yield at most $1+20(8-1)=141$ representative intervals rather than $8^{20}$ combinations. This is a bound on \emph{envelope intervals}, not on the cost of generating the candidate pools or computing the majorizer.

\begin{theorem}[Surrogate optimality within the candidate pools]
\label{thm:pool_opt}
Let each $\mathcal P_i$ be finite and nonempty, containing the current index, and let $\mathcal S\subset(0,\infty)$ be either a compact real interval or a nonempty finite set of legally representable scales within such an interval. If the lower envelopes are constructed exactly and every interval is minimized over its feasible scales with correct boundary handling, the interval procedure returns a global minimizer of $Q(k,s\mid z_0)$ on $\bigl(\prod_i\mathcal P_i\bigr)\times\mathcal S$.
\end{theorem}

\paragraph{Proof.}
Partition the admissible continuous scale interval by all distinct interior envelope breakpoints. On each nonempty open subinterval, a minimizing codeword index can be chosen constant at every position; denote the joint choice by $k^{(r)}$. At any breakpoint, the limiting representatives from adjacent intervals both attain $L_i(s)$ at every position they change, so at least one adjacent joint representative realizes the exact envelope minimum. A line that is minimal only at an isolated breakpoint does not need a separate joint combination for the \emph{surrogate value}, because the neighboring envelope lines tie with it there. It follows that minimizing the quadratic $Q(k^{(r)},s\mid z_0)$ over the closure of each interval and taking the best value recovers the minimum of Eq.~\ref{eq:app_min_envelope} over the full scale interval.

For continuous $s$ and $A_r>0$, $Q_r(s)=C+A_rs^2-2B_rs$ is convex and its unique unconstrained minimizer is $B_r/A_r$, giving Eq.~\ref{eq:sr} after clipping. If $A_r=0$, positive definiteness of $D_i$ implies that all selected $c_{ik_i^{(r)}}=0$, and therefore also $B_r=0$; the surrogate is then constant over the interval. More general implementations allowing semidefinite or approximate coefficients may handle $A_r=0$ as a linear function, choosing a legal endpoint according to the sign of $B_r$.

For a finite representable scale set, restrict each interval closure to its legal elements. Convexity implies that a minimum occurs at a legal predecessor or successor surrounding the continuous vertex, or at the nearest legal endpoint if the vertex lies outside the interval. Breakpoint values that are exactly representable are included explicitly; candidate combinations from neighboring intervals remain optimal there. Empty intersections are skipped. When the domain consists of a single scale, directly select the lowest-cost codeword at every position. In all cases, every legal scale is considered through at least one interval closure or this singleton rule. Comparing all such minima therefore gives the global surrogate minimum over the specified pools and legal scales. \hfill$\square$

\paragraph{What the theorem does not claim.}
The theorem neither certifies the full-codebook optimum (because $\mathcal P_i$ may be truncated) nor the true-objective optimum (because $Q$ replaces cross-position curvature by a majorizer). On envelope boundaries, different tied assignments may have different \emph{true} errors even though their surrogate values coincide. The basic solver selects one deterministic surrogate-optimal assignment and verifies that proposal, without enumerating all tied true-objective alternatives.

\subsubsection{True-Error Decrease, Surrogate Gap, and Acceptance}
\label{app:acceptance_proof}
\paragraph{Non-increase and additive gap.}
Let $z_Q$ minimize $Q$ over the selected candidate pools and legal-scale set. The current stored state $z_0$ is feasible by construction. Proposition~\ref{prop:majorizer} therefore yields
\begin{equation}
F(z_Q)\le Q(z_Q\mid z_0)
\le Q(z_0\mid z_0)=F(z_0).
\label{eq:app_nonincrease}
\end{equation}
This is a monotonicity guarantee \emph{before} the strict acceptance threshold, assuming exact surrogate minimization and consistent evaluation of the stored reconstruction. The actual algorithm additionally requires $F(z_Q)\le F(z_0)-\eta$. In particular, $Q(z_Q\mid z_0)\le F(z_0)-\eta$ is a sufficient certificate for acceptance but is not necessary: the upper bound may be conservative, so a move that improves the true objective by $\eta$ can fail the stronger surrogate test.

Let $z_F$ be a minimizer of the true objective $F$ \emph{over the same pools and legal scales}, and write $\Delta_F=z_F-z_0$. By Eq.~\ref{eq:gap},
\begin{equation}
\begin{aligned}
F(z_Q)&\le Q(z_Q\mid z_0)\le Q(z_F\mid z_0)\\
&=F(z_F)+\Delta_F^\top(D-H)\Delta_F.
\end{aligned}
\label{eq:QF}
\end{equation}
As $F(z_F)\le F(z_Q)$, we obtain
\begin{equation}
\begin{aligned}
0\le {}&F(z_Q)-F(z_F)\\
\le {}&(z_F-z_0)^\top(D-H)(z_F-z_0)\\
\le {}&\|D-H\|_2\|z_F-z_0\|_2^2.
\end{aligned}
\label{eq:gap_bound}
\end{equation}
This bounds the \emph{additive} surrogate-induced suboptimality relative to the true minimizer in the \emph{same pools}, not the entire unquantized parameter space. If $D=H$ and $H$ is block diagonal, $Q\equiv F$ on the group and solver optimality transfers directly to that group's true objective. The size of the bound depends on the majorizer slack; a valid but loose majorizer may still miss true-improving joint moves. Thus the coordinated-update example in Appendix~\ref{app:example} establishes the existence of coordinate-wise stalls, not an unconditional escape theorem for \method{}.

\paragraph{Actual storage and finite termination.}
The accepted proposal is evaluated after packing the chosen scale and hard indices into their deployed representation. Rounding should not be applied \emph{after} checking the true error. If all feasible positive scales are representable in a finite storage dtype, the complete stored-state space is finite. With $\eta>0$, every accepted move decreases the nonnegative true error by at least $\eta$, giving no more than $\lfloor F(z^{(0)})/\eta\rfloor$ accepted moves when $\eta$ is fixed. In exact continuous-scale optimization, finiteness of hard-index combinations alone does not prove finite termination, because the scale is continuous; the positive absolute threshold or an explicit sweep limit supplies a separate stopping argument.

\subsubsection{Conditions for Preserving Envelope Combinations}
\label{app:preserved_combinations}
This section characterizes when replacing the block-diagonal curvature $H_{\mathrm{diag}}=\operatorname{blkdiag}(H_{11},\ldots,H_{MM})$ by $D$ leaves candidate choices unchanged. Define $D_i=H_{ii}+\lambda_i I_d$ with $\lambda_i\ge0$. For the comparison built with $H_{ii}$, let $a_{ik}^{H}=c_{ik}^\top H_{ii}c_{ik}$ and $b_{ik}^{H}=c_{ik}^\top(H_{ii}z_{0i}-g_i)$. Let $a_{ik}^{D},b_{ik}^{D}$ denote the corresponding quantities under $D_i$. Then
\begin{equation}
\begin{aligned}
a_{ik}^{D}-a_{ik}^{H}&=\lambda_i\|c_{ik}\|_2^2,\\
b_{ik}^{D}-b_{ik}^{H}&=\lambda_i c_{ik}^\top z_{0i}.
\end{aligned}
\label{eq:loaded_coeff_differences}
\end{equation}
Accordingly, the pairwise difference between candidate $k$ and $k'$ is shifted by
\begin{equation}
\begin{aligned}
&\bigl[\ell_{ik}^{D}(s)-\ell_{ik'}^{D}(s)\bigr]
 -\bigl[\ell_{ik}^{H}(s)-\ell_{ik'}^{H}(s)\bigr]\\
&\quad=\lambda_i\bigl[s(\|c_{ik}\|_2^2-\|c_{ik'}\|_2^2)\\
&\qquad\qquad-2(c_{ik}-c_{ik'})^\top z_{0i}\bigr].
\end{aligned}
\label{eq:loaded_line_difference}
\end{equation}
A sufficient condition for identical pointwise minimizing assignments is that, for every position, candidate pair, and $s\in\mathcal S$, the corresponding pairwise line differences under $D$ and under $H_{\mathrm{diag}}$ have the same sign, including ties. A stronger and easily checked sufficient condition is that the curvature-loading term in Eq.~\ref{eq:loaded_line_difference} vanish identically for every candidate pair. For $\lambda_i>0$, this holds if all candidates at position $i$ have both identical squared norms and identical inner products with $z_{0i}$. \emph{Equal codeword norms alone are not sufficient}: the second, linear-in-$z_{0i}$ term can still alter rankings. If the sign condition holds for every position, the pointwise argmin sets and resulting interval representatives are preserved (up to deterministic tie selection); even then, the minimizing scale within an interval can change because its quadratic coefficients differ. These preservation conditions concern surrogate candidate ordering and do not restore the omitted off-diagonal interactions of the true objective.

\subsection{Implementation and Fair-Comparison Details}
\label{app:impl}

\subsubsection{Actual Bitrate and Representation Scope}
\label{app:impl:rate}
The actual effective bits per weight should include every stored component:
\begin{equation}
\mathrm{bpw}=\frac{B_{\mathrm{indices}}+B_{\mathrm{scales}}+
B_{\mathrm{codebooks}}+B_{\mathrm{aux}}}{N_{\mathrm{weights}}}.
\label{eq:actual_bpw}
\end{equation}
Fixed codebooks still occupy storage unless they are a specified external, shared, or analytically generated resource. Index packing, group-scale metadata, alignment, tails, and non-negligible auxiliary parameters must be counted consistently. When lossless coding is used for the compressed checkpoint, report both the effective representation rate and the actual serialized checkpoint size, as they capture different costs. Tail handling for $d=6$ must write back only legal parameter positions; padded vector entries must not be treated as real weights in $\mathcal A$, $H$, or the error objective.

\subsubsection{Group Scope, Numerical Precision, and Verification}
\label{app:impl:scale}
Group updates include \emph{every} position covered by the scale, even if only $m$ indices can change. Singleton positions contribute quadratic coefficients but no switching points. To study the effect of $m$, keep the existing scale-group partition, codebooks, and scale count fixed; otherwise changes in scale granularity confound the comparison. If weights are reconstructed in a rotated coordinate system, both $W$ and $X$ must be defined in the corresponding transformed coordinates and decoding must consistently apply any inverse transform.

For discrete scales, record the stored dtype, positive admissible range, and policy for subnormal and boundary values. Higher precision may be used to compute intersection points and coefficients, but actual objective verification must use the reconstructed weights \emph{after} converting the chosen scale to the stored dtype. When the nearest representable scale lies outside a candidate interval, use the inward predecessor or successor rather than associating that rounded value with an invalid interval representative. The stopping rule should state the exact absolute and relative thresholds used; when both are used, a defensible convention is $\eta_t=\max\{\eta_{\mathrm{abs}},\eta_{\mathrm{rel}}F(z_0)\}$ with $\eta_{\mathrm{abs}}>0$. This specifies a reproducible policy but should match the actual implementation, rather than being presented as an experimental setting without verification.

\subsubsection{Reproducibility and Fair Evaluation}
\label{app:impl:repro}
Implementations can batch-evaluate candidate coefficients before building position-wise envelopes. For memory efficiency, record the initial codeword representative, the sequence of switching events, and the best interval ID; replay the events once to reconstruct the best full index assignment instead of storing one $M$-length assignment per interval. Only the final surrogate-minimizing proposal is verified by the true error in the base algorithm. Optional reranking of every interval representative or gradient-based post-search scale refinement should be reported as a separate algorithmic step with its additional cost.

Experimental reporting should fix configurations, seeds, model-checkpoint identifiers, codebook construction, calibration samples, and software versions. Accuracy improvements on calibration data must not be substituted for unseen benchmark results. Comparisons to numbers published elsewhere should identify their source and any differences in quantized layers, actual bit budgets, calibration data, or evaluation protocol. If a single-GPU sequential execution policy is used, record it alongside measured preprocessing, search, fine-tuning, and inference runtimes so that envelope-only complexity is not mistaken for total wall-clock cost.

\subsection{A Deterministic Lower-Envelope Example}
\label{app:figures}

Consider two positions with candidate comparison lines
\begin{equation}
\begin{gathered}
\ell_A(s)=3s,\quad \ell_B(s)=2s+1,\quad \ell_E(s)=s+3,\\
\ell_C(s)=2s,\qquad\ell_D(s)=s+\tfrac32.
\end{gathered}
\label{eq:app_example_lines}
\end{equation}
The first three lines belong to position 1; the last two belong to position 2. Their nonnegative slopes are consistent with $a_{ik}=c_{ik}^\top D_i c_{ik}\ge0$, and the intercepts equal $-2b_{ik}$. Position 1 changes its lower-envelope minimizer from $A$ to $B$ at $s=1$, then from $B$ to $E$ at $s=2$. Position 2 changes from $C$ to $D$ at $s=3/2$. On the legal interval $\mathcal S=[1/4,3]$, merging the switching points gives
\begin{equation}
\begin{gathered}
\left[\tfrac14,1\right],\quad
\left[1,\tfrac32\right],\quad
\left[\tfrac32,2\right],\quad[2,3],\\
(A,C),\quad(B,C),\quad(B,D),\quad(E,D).
\end{gathered}
\label{eq:app_interval_example}
\end{equation}
The second line lists the respective joint representatives. At a shared endpoint, the adjacent representatives tie in the surrogate, so either may be selected consistently. There are four intervals, meeting the bound $1+(3-1)+(2-1)=4$ in Eq.~\ref{eq:R}. To obtain the complete surrogate optimum, each representative must still be evaluated using its \emph{quadratic} $C+A_rs^2-2B_rs$ over the corresponding interval; choosing a minimum of the \emph{linear} envelopes alone would be incorrect. The example illustrates the envelope geometry, not a particular prescribed value of the final optimal scale or a guarantee about the true layer-output error.
% 附录：补充结果
\subsection{Supplementary Results}
\label{app:all-results}

This appendix records the per-model task breakdown that complements the compact main tables. In Table~\ref{tab:main-ppl}, PPL.AVG is the arithmetic mean of WikiText-2 and C4 perplexities, and ACC.AVG is the macro-average of ARC-Challenge, ARC-Easy, HellaSwag, LAMBADA OpenAI, PIQA, and WinoGrande~\cite{arc,hellaswag,lambada,piqa,winogrande}. The component perplexities and remaining zero-shot accuracies appear in the per-model tables below~\cite{wikitext,c4,boolq,mmlu,ifeval}. IFEval is reported only for instruction models and uses prompt-level strict accuracy. Figure~\ref{fig:qwen3-4b-overview} in the main text visualizes the Qwen3-4B comparison.

\begin{table*}[!bp]
\caption{Per-task results on Qwen3-4B. WikiText-2 and C4 are perplexities (lower is better); the remaining columns are zero-shot accuracies (\%, higher is better). Bold denotes the best result among quantized methods.}
\label{tab:app-qwen3-4b}
\centering
\small
\setlength{\tabcolsep}{3.0pt}
\resizebox{\linewidth}{!}{%
\begin{tabular}{lcccccccccc}
\toprule
Method & Wiki2$\downarrow$ & C4$\downarrow$ & ARC-C$\uparrow$ & ARC-E$\uparrow$ & HellaS$\uparrow$ & LAMBADA$\uparrow$ & PIQA$\uparrow$ & BoolQ$\uparrow$ & MMLU$\uparrow$ & WinoG$\uparrow$ \\
\midrule
BF16 & 13.66 & 17.88 & 53.67 & 78.49 & 68.43 & 59.40 & 74.97 & 84.98 & 68.22 & 65.75 \\
\midrule
PV-Tuning & 14.87 & 19.33 & 43.52 & 69.49 & 60.58 & 50.46 & 72.91 & 77.55 & 55.10 & 60.46 \\
QuIP\# & 17.98 & 24.98 & 40.27 & 66.58 & 56.94 & 47.91 & 70.08 & 77.92 & 52.90 & 60.38 \\
QTIP & 15.78 & 20.62 & \textbf{43.77} & 68.69 & 60.85 & 51.58 & 73.12 & \textbf{84.13} & \textbf{58.58} & \textbf{63.77} \\
NoWag & 22.69 & 24.22 & 42.24 & 65.87 & 55.73 & 46.32 & 71.00 & 79.14 & 53.49 & 61.96 \\
AQLM & 15.12 & 23.38 & 40.96 & 67.89 & 62.92 & 55.19 & 73.07 & 78.32 & 56.11 & 60.46 \\
LLVQ & 14.27 & 17.99 & 42.58 & \textbf{71.30} & \textbf{63.48} & 54.22 & 72.09 & 77.00 & 57.80 & 60.93 \\
\method{} & \textbf{13.68} & \textbf{17.97} & 42.92 & 70.12 & 63.16 & \textbf{55.79} & \textbf{73.61} & 77.13 & 55.01 & 62.90 \\
\bottomrule
\end{tabular}%
}

\end{table*}

\begin{table*}[!bp]
\caption{Per-task results on Qwen3-8B. WikiText-2 and C4 are perplexities (lower is better); the remaining columns are zero-shot accuracies (\%, higher is better). Bold denotes the best result among quantized methods.}
\label{tab:app-qwen3-8b}
\centering
\small
\setlength{\tabcolsep}{3.0pt}
\resizebox{\linewidth}{!}{%
\begin{tabular}{lcccccccccc}
\toprule
Method & Wiki2$\downarrow$ & C4$\downarrow$ & ARC-C$\uparrow$ & ARC-E$\uparrow$ & HellaS$\uparrow$ & LAMBADA$\uparrow$ & PIQA$\uparrow$ & BoolQ$\uparrow$ & MMLU$\uparrow$ & WinoG$\uparrow$ \\
\midrule
BF16 & 9.72 & 14.29 & 56.74 & 80.68 & 74.94 & 64.08 & 77.69 & 86.61 & 72.97 & 68.11 \\
\midrule
PV-Tuning & 10.51 & 15.28 & 51.45 & 77.36 & 68.02 & 61.67 & 75.90 & 81.41 & 63.52 & 66.61 \\
QuIP\# & 12.64 & 18.88 & 49.83 & 76.30 & 64.82 & 60.99 & 73.45 & 81.68 & 63.70 & 66.54 \\
QTIP & 11.61 & 16.20 & 50.77 & 75.42 & 68.48 & 62.68 & \textbf{76.33} & \textbf{84.50} & 66.03 & 67.09 \\
NoWag & 13.85 & 18.07 & 51.79 & 78.20 & 63.41 & 59.38 & 73.72 & 83.36 & 62.38 & 67.88 \\
AQLM & 10.71 & 17.84 & 50.17 & 76.81 & 70.09 & \textbf{62.95} & 76.06 & 81.96 & 64.25 & 66.61 \\
LLVQ & \textbf{8.78} & \textbf{14.36} & 50.94 & 78.03 & \textbf{70.58} & 62.68 & 75.19 & 80.95 & \textbf{69.60} & 66.69 \\
\method{} & 9.76 & 14.64 & \textbf{54.52} & \textbf{79.92} & 69.80 & 62.74 & 76.01 & 80.55 & 63.23 & \textbf{68.67} \\
\bottomrule
\end{tabular}%
}

\end{table*}

\begin{table*}[!bp]
\caption{Per-task results on Qwen3-14B. WikiText-2 and C4 are perplexities (lower is better); the remaining columns are zero-shot accuracies (\%, higher is better). Bold denotes the best result among quantized methods.}
\label{tab:app-qwen3-14b}
\centering
\small
\setlength{\tabcolsep}{3.0pt}
\resizebox{\linewidth}{!}{%
\begin{tabular}{lcccccccccc}
\toprule
Method & Wiki2$\downarrow$ & C4$\downarrow$ & ARC-C$\uparrow$ & ARC-E$\uparrow$ & HellaS$\uparrow$ & LAMBADA$\uparrow$ & PIQA$\uparrow$ & BoolQ$\uparrow$ & MMLU$\uparrow$ & WinoG$\uparrow$ \\
\midrule
BF16 & 8.65 & 12.85 & 60.24 & 82.83 & 78.86 & 67.88 & 79.87 & 89.33 & 77.20 & 72.93 \\
\midrule
PV-Tuning & 9.15 & 13.56 & 55.46 & 79.71 & 73.60 & 66.54 & 78.24 & 83.82 & 69.30 & 71.98 \\
QuIP\# & 10.43 & 16.31 & 53.92 & 78.70 & 71.16 & 66.16 & 76.06 & 84.10 & 67.98 & 71.90 \\
QTIP & 9.84 & 14.33 & \textbf{56.57} & 78.70 & 73.86 & \textbf{68.97} & 78.18 & \textbf{88.38} & \textbf{71.17} & \textbf{72.69} \\
NoWag & 11.04 & 15.60 & 53.67 & 79.42 & 70.23 & 65.19 & 77.26 & 87.19 & 68.69 & 71.98 \\
AQLM & 9.25 & 15.53 & 54.27 & 79.17 & 75.17 & 67.24 & 78.40 & 84.40 & 69.91 & 71.90 \\
LLVQ & 8.98 & 13.66 & 54.95 & \textbf{80.35} & \textbf{75.54} & 67.11 & 77.64 & 83.43 & 70.93 & 72.06 \\
\method{} & \textbf{8.51} & \textbf{13.09} & 53.24 & 79.21 & 75.30 & 67.36 & \textbf{78.94} & 86.73 & 69.76 & 71.98 \\
\bottomrule
\end{tabular}%
}

\end{table*}

\begin{table*}[!bp]
\caption{Per-task results on Qwen3-32B. WikiText-2 and C4 are perplexities (lower is better); the remaining columns are zero-shot accuracies (\%, higher is better). Bold denotes the best result among quantized methods.}
\label{tab:app-qwen3-32b}
\centering
\small
\setlength{\tabcolsep}{3.0pt}
\resizebox{\linewidth}{!}{%
\begin{tabular}{lcccccccccc}
\toprule
Method & Wiki2$\downarrow$ & C4$\downarrow$ & ARC-C$\uparrow$ & ARC-E$\uparrow$ & HellaS$\uparrow$ & LAMBADA$\uparrow$ & PIQA$\uparrow$ & BoolQ$\uparrow$ & MMLU$\uparrow$ & WinoG$\uparrow$ \\
\midrule
BF16 & 7.61 & 11.51 & 60.92 & 83.16 & 82.58 & 67.18 & 82.10 & 86.36 & 80.86 & 72.93 \\
\midrule
PV-Tuning & 8.10 & 12.26 & 60.49 & 82.70 & 78.47 & 70.46 & 79.43 & 86.36 & 74.62 & 75.53 \\
QuIP\# & 9.73 & 14.57 & 60.58 & 81.44 & 76.67 & 69.47 & 77.75 & 85.47 & 72.22 & 74.59 \\
QTIP & 8.53 & 12.49 & 60.24 & 81.69 & \textbf{79.74} & 69.77 & 79.60 & \textbf{88.29} & \textbf{77.28} & 74.19 \\
NoWag & 9.19 & 13.35 & \textbf{61.12} & \textbf{83.08} & 77.03 & 70.60 & 79.00 & 88.13 & 74.13 & 74.90 \\
AQLM & 8.15 & 13.20 & \textbf{61.12} & 82.53 & 78.97 & 70.42 & 79.49 & 86.61 & 74.77 & 75.22 \\
LLVQ & 7.87 & 12.15 & 61.09 & 82.91 & 79.09 & 70.42 & 79.22 & 86.18 & 76.35 & 75.45 \\
\method{} & \textbf{7.75} & \textbf{11.99} & 60.32 & 82.74 & 78.17 & \textbf{70.83} & \textbf{79.84} & 86.94 & 76.91 & \textbf{76.09} \\
\bottomrule
\end{tabular}%
}

\end{table*}

\begin{table*}[!bp]
\caption{Per-task results on Llama-3.1-8B-Instruct. WikiText-2 and C4 are perplexities (lower is better); the remaining columns are zero-shot accuracies (\%, higher is better). Bold denotes the best result among quantized methods.}
\label{tab:app-llama31-8b}
\centering
\small
\setlength{\tabcolsep}{3.0pt}
\resizebox{\linewidth}{!}{%
\begin{tabular}{lccccccccccc}
\toprule
Method & Wiki2$\downarrow$ & C4$\downarrow$ & ARC-C$\uparrow$ & ARC-E$\uparrow$ & HellaS$\uparrow$ & LAMBADA$\uparrow$ & PIQA$\uparrow$ & BoolQ$\uparrow$ & MMLU$\uparrow$ & WinoG$\uparrow$ & IFEval$\uparrow$ \\
\midrule
BF16 & 7.22 & 10.63 & 55.03 & 79.67 & 79.17 & 73.08 & 80.96 & 83.85 & 67.93 & 73.40 & 73.94 \\
\midrule
PV-Tuning & 8.45 & 13.10 & 46.16 & 75.42 & 54.40 & 64.10 & \textbf{78.73} & 76.39 & 54.80 & \textbf{70.56} & 60.44 \\
QuIP\# & 9.75 & 14.85 & 43.26 & 69.65 & 67.70 & 61.60 & 76.82 & 76.79 & 52.60 & 69.22 & 55.08 \\
QTIP & 8.80 & 13.39 & \textbf{50.77} & 74.34 & \textbf{75.02} & 69.71 & 78.24 & 80.01 & 54.39 & 69.01 & \textbf{64.25} \\
NoWag & 10.50 & 15.28 & 47.27 & 74.09 & 70.14 & 67.48 & 77.09 & \textbf{80.52} & 55.95 & 69.93 & 56.75 \\
AQLM & 9.50 & 15.00 & 30.89 & 46.25 & 61.40 & 60.00 & 76.50 & 68.01 & 42.29 & 58.25 & 40.48 \\
LLVQ & \textbf{8.27} & 13.44 & 48.29 & 75.17 & 73.76 & 67.81 & 77.48 & 78.53 & \textbf{57.67} & 70.01 & 61.55 \\
\method{} & 8.38 & \textbf{12.74} & 47.95 & \textbf{75.51} & 74.09 & \textbf{69.74} & 77.37 & 79.77 & 55.16 & 70.32 & 60.46 \\
\bottomrule
\end{tabular}%
}

\end{table*}

\begin{table*}[!bp]
\caption{Per-task results on Llama-3.1-70B-Instruct. WikiText-2 and C4 are perplexities (lower is better); the remaining columns are zero-shot accuracies (\%, higher is better). Bold denotes the best result among quantized methods.}
\label{tab:app-llama31-70b}
\centering
\small
\setlength{\tabcolsep}{3.0pt}
\resizebox{\linewidth}{!}{%
\begin{tabular}{lccccccccccc}
\toprule
Method & Wiki2$\downarrow$ & C4$\downarrow$ & ARC-C$\uparrow$ & ARC-E$\uparrow$ & HellaS$\uparrow$ & LAMBADA$\uparrow$ & PIQA$\uparrow$ & BoolQ$\uparrow$ & MMLU$\uparrow$ & WinoG$\uparrow$ & IFEval$\uparrow$ \\
\midrule
BF16 & 3.78 & 8.00 & 63.48 & 83.84 & 84.63 & 75.68 & 83.79 & 87.68 & 82.37 & 78.77 & 83.55 \\
\midrule
PV-Tuning & 5.70 & 10.95 & 52.73 & 82.20 & 76.25 & 72.00 & \textbf{82.59} & 84.71 & 72.50 & 75.45 & 73.57 \\
QuIP\# & 5.56 & 11.21 & 50.68 & 79.76 & 72.74 & 68.60 & 81.77 & 83.03 & 70.89 & 75.14 & 70.61 \\
QTIP & 5.08 & \textbf{10.48} & 54.40 & 82.60 & \textbf{76.40} & \textbf{72.65} & 82.50 & 85.12 & \textbf{73.78} & 75.51 & \textbf{75.73} \\
NoWag & 5.93 & 11.43 & 52.65 & \textbf{83.00} & 73.96 & 71.53 & 81.94 & 84.89 & 72.56 & 75.45 & 71.53 \\
AQLM & 5.33 & 11.04 & 47.44 & 72.43 & 71.63 & 69.24 & 81.88 & 80.58 & 68.15 & 71.43 & 67.10 \\
LLVQ & \textbf{4.82} & 10.71 & 53.16 & 82.53 & 75.77 & 71.71 & 82.10 & 83.88 & 73.42 & 75.53 & 73.94 \\
\method{} & \textbf{4.82} & 10.63 & \textbf{54.99} & 82.70 & 75.94 & 72.46 & 82.55 & \textbf{85.49} & 73.57 & \textbf{75.69} & 68.39 \\
\bottomrule
\end{tabular}%
}

\end{table*}

\end{document}